# Comprehensive Forecasting-Based Analysis of Hybrid and Stacked Stateful/ Stateless Models

*Note: Sub-titles are not captured in Xplore and should not be used

Swayamjit Saha
*Dept. of Computer Science and Engineering*
*Mississippi State University*
Starkville, United States of America
ss4706@msstate.edu

*Abstract*—Wind speed is a powerful source of renewable energy, which can be used as an alternative to the non-renewable resources for production of electricity. Renewable sources are clean, infinite and do not impact the environment negatively during production of electrical energy. However, while eliciting electrical energy from renewable resources viz. solar irradiance, wind speed, hydro should require special planning failing which may result in huge loss of labour and money for setting up the system. In this paper, we discuss four deep recurrent neural networks viz. Stacked Stateless LSTM, Stacked Stateless GRU, Stacked Stateful LSTM and Statcked Stateful GRU which will be used to predict wind speed on a short-term basis for the airport sites beside two campuses of Mississippi State University. The paper does a comprehensive analysis of the performance of the models used describing their architectures and how efficiently they elicit the results with the help of RMSE values. A detailed description of the time and space complexities of the above models has also been discussed.

## I. MOTIVATION

To find the best fit hybrid stacked architecture deep neural network in terms of wind speed prediction To find an accurate location for installation of wind turbine so that we can attain the maximum sustainable energy output from that location.

## II. INTRODUCTION

Before delving into this project details, there is need to address some motivating factors that brought about this project idea realization. The approach towards the exploration of Energy Generation and Production without first addressing some critical issues will be incomplete. Since the inception of power generation, humankind has been faced with the constant problem of mitigating climate change effects like the greenhouse effect. However, let us first describe the term "Energy Generation." In the layman's term, Energy Generation is the process of producing energy needed for humankind related activities. According to Wikipedia, Energy Generation is the process of generating power from sources of primary energy [1]. Before Electricity can be produced, Energy must be generated first, and then transformed to the most usual form of energy known to man as electricity.Recurrent Neural Networks

1.2. Classification of Energy

Energy sources are classified into two broad categories which are.

- Renewable Energy
- Non-Renewable Energy

Renewable Energy comes from natural processes replenished constantly such as sunlight, wind, ocean, hydropower, biomass, geothermal resources, and biofuels and hydrogen. It is interesting to know that they are replenished at a higher rate than they are consumed [2]. Its advantages include the following.

- It is non-polluting
- It does not emit any greenhouse gases
- It is available in plenty

Non-Renewable Energy is derived from sources that will not be replenished in a lifetime. Its major source is fossil fuels, which includes coal, petroleum, and natural gas. It is interesting to know that they are the largest sources of energy for electricity generation. Based on stats from the US Energy Information Administration, Non-renewables accounts for up to 81% of the total electricity generation in the US in 2021, with natural gas being the most common source (38%) [3].

Generating renewable energy creates far lower emissions than burning fossil fuels. Transitioning from fossil fuels, which currently account for the lion's share of emissions, to renewable energy is key to addressing the climate crisis [2].

1.3. Climate Change

Climate Change refers to long-term shifts in temperatures and weather patterns. Since the 1800s, human activities have been the major driver of climate change, primarily due to burning fossil fuels like coal, oil, and gas, which in turn generates gas emissions [2].

The greenhouse effect is essential to life on Earth, but human-made emissions in the atmosphere are trapping and slowing heal loss to space. Five key greenhouse gases are CO2, nitrous oxide, methane, chlorofluorocarbons, and water vapor [4]. The consequences of climate change include intense droughts, water scarcity, severe fires, rising sea levels, flooding, melting polar ice, catastrophic storms, and declining biodiversity.

### 1.4. Machine Learning in predicting wind speed

Wind Energy is simply energy generated by using wind to spin a wind turbine to generate electricity. The kinetic energy in the airflow is converted into electrical energy by wind turbines. Assessing the characteristics and potential of wind energy is the first step in the effective development of wind energy. Some of the drawbacks of utilizing wind as a source for energy generation includes the expensive up-front costs, potential to endanger wildlife, and unpredictable wind behavior.

Though energy generation through the wind brings so much promise to the environment and to human activities, challenges such as locating ideal sites for the installation of wind turbines need to be addressed. Upgrading the nation's transmission network to connect areas with abundant wind resources to population centers could significantly reduce the costs of expanding land-based wind energy.

Predicting wind speed and direction is one of the most crucial and critical tasks in a wind farm because wind turbine blades motion and energy production is closely related to the wind flow's behavior. In this project, we will be doing a comparative analysis of different Recurrent Neural Networks (RNNs) to find the best fit model for accurate forecasting.

### 1.5. Recurrent Neural Networks

RNN (Recurrent Neural Networks) is a class of artificial neural networks where connections between nodes can create a cycle, allowing output from some nodes to affect subsequent input to the same nodes [6]. This makes it suitable for sequential data or time series data. It is good for time-series data because it consists of the Hidden state, which remembers some information about a sequence. However, the major drawbacks of this network are exploding and vanishing gradients.

RNN has distinct types that help mitigate this problem. Such types include Long Short-term Memory (LSTM) and Gated Recurrent Units (GRUs). With time series data, LSTM networks are well suited for classifying, processing, and making predictions based on data. It is comprised of a cell, an input gate, an output gate, and a forget gate. GRUs are also provided with a gated mechanism to capture dependencies of different time scales effectively and adaptively. In contrast to LSTMs, GRUs do not store cell states hence they are unable to regulate the amount of memory content. They also exhibit better performance on certain smaller and less frequent datasets. In the results section of this report, we present the graphic representation of the results gotten for both LSTM and GRU for the stacked stateful and stacked stateless models.

### 1.6. Related work

A lot of work has been done in recent years regarding wind prediction. The stochastic nature of wind makes it difficult to develop models which accurately predict future wind speed and direction. As a result, many different approaches have been tried to find the ideal prediction model.

The concept of comparing Back Propagation Network, Radial Basis Function, and Nonlinear Autoregressive model process with exogenous inputs (NARX) with Mutual Information feature (MI)selection for wind speed prediction was introduced in [6]. NARX with MI was found to have outperformed other models based on the Root Mean Square Error (RMSE) and Mean Absolute Error (MAE) values.

In [7], two models were proposed individually for prediction of wind direction and prediction of wind speed. An ANN model was trained and such algorithms such as Random Forest algorithm (wind direction) and Linear Regression algorithm (wind speed) were also being implemented. These algorithms were evaluated based on a ten-minute interval measurement. The proposed models are said to be used for non-extreme weather conditions when there is no turbulence.

### 1.7. Problem Introduction

We plan to compare a set of four Deep Learning models with varied stacked architectures to predict wind speed in four remote locations across Mississippi State University campuses situated in Starkville and Meridian. We plan to use publicly available time series data from the Surface Radiation Budget Network (SURFRAD) database (https://gml.noaa.gov/grad/surfrad/) to extract the most key features for prediction, calculate the accuracy of each model, then conclude which deep learning model performs the best in terms of prediction of wind speed as opposed to the other models.

### *C. Gated Recurrent Units (GRUs)*

A Gated Recurrent Unit (GRU) introduced by Cho, K. et al. in their paper [4], is a variation of RNN architecture which employs gates to control the flow of information between the various cells of an unit. GRU is a new model as compared to LSTM architecture introduced by Hochreiter, S. et al in their paper. [9] The ability of GRU to adhere to long-term dependence or memory from the internal direction of the GRU cell to produce a hidden state. While LSTMs have two different states that have passed between cells - cell status and hidden state, with long and short memory, respectively - only GRUs have one hidden state transferred between time intervals. This hidden state is capable of capturing long-term and short-term reliability due to hidden processing methods and accounting and input data. The GRU cell contains only two gates: the Update gate and the Reset gate. Like gates on LSTMs, these GRU gates are trained to filter out any inaccurate information while maintaining usability. These gates are basically vans containing 0 to 1 values that will be expanded with input data and / or hidden country. A value of 0 on the gate vases indicates that the corresponding data in the installation or hidden region is insignificant and, therefore, will return as zero. On the other hand, the number 1 in the image at the gate means that the corresponding data is important and will be used. In this paper the terms gate and vector are synonymous and will be used interchangeably. The structure of a GRU unit is shown in Figure 1.

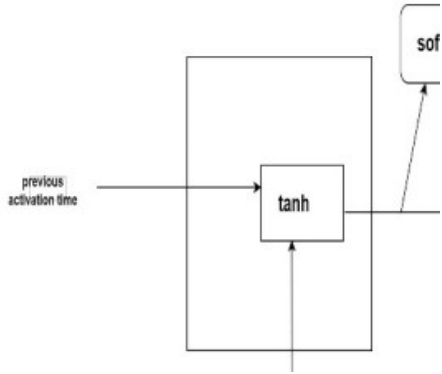

The output can be calculated by including the inputs of previous activation time step and , times. Now we consider a new variable for memory cell cm. Let us consider a sample statement: The child, which already ate… was full.

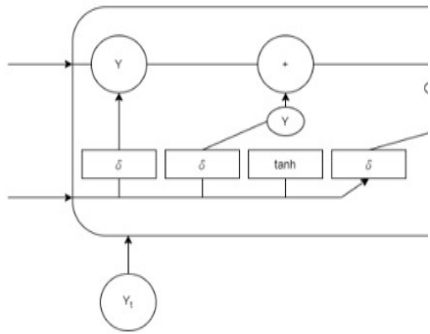

**Fig 2 A Single Unit of LSTM cell**

### III. METHODOLOGY

We will use publicly available hourly time-series wind speed data from the SURFRAD database to extract meteorological features of specific locations across the two MSU campuses. We then plan to use correlation analysis to identify the features in the dataset that are most important to predicting wind speed output. Once we have identified our training features, we will compare four different stacked deep learning models: Stacked Stateful and Stacked Stateless Long Short-Term Memory (LTSM), and Stacked Stateful and Stacked Stateless Gated Recurrent Unit network (GRU). We will perform short-term wind speed forecasting by training our models on three months of data and predicting one to two days of the above locations. We will evaluate the models' accuracies by calculating root mean squared error (RMSE), mean squared error (MSE) and F1 scores. Finally, we will do a comparative performance-based analysis to estimate which model is best suited for prediction of wind speed among the rest considered models. The parameters used are Hourly Dewpoint Temperature, Hourly Dry Bulb Temperature, Hourly Relative Humidity, Hourly Sea Level Pressure, Hourly Station Pressure, Hourly visibility, Hourly Wet Bulb Temperature, Hourly Wind Direction, and Hourly Wind Speed.

Data Set Extraction: The dataset is extracted from the National centers for Environmental Information National Oceanic and Atmospheric Administration (NOAA) (https://www.ncdc.noaa.gov/) which is publicly available.

Data Optimization: Raw data consists of iterative parameters were ignored for our experiment. We considered forecasted parameters viz. Hourly Dewpoint Temperature, Hourly Dry Bulb Temperature, Hourly Relative Humidity, Hourly Sea Level Pressure, Hourly Station Pressure, Hourly visibility, Hourly Wet Bulb Temperature, Hourly Wind Direction, and Hourly Wind Speed to get accurate predictions.

Neural Networks: The optimized data is trained with neural networks. The selected recurrent neural networks are: stacked stateless LSTM (SSLL- STM), stacked stateful LSTM (SSFLSTM), stacked stateless GRU (SSLGRU), stacked stateful GRU (SSFGRU).

Result analysis: The results from the four neural network architectures are collected and then compared. This is done to show which model performs better for forecasting solar irradiance.

Conclusion: This stage determines which model performs the best among the four models chosen. Finally, we present the algorithm for the best-proven model for forecasting wind speed.

Data was collected from the National centers for Environmental Information National Oceanic and Atmospheric Administration (NOAA) (https://www.ncdc.noaa.gov/) public database. We collected data from two different locations that were a proxy to the two different Mississippi State University campuses in Meridian (Meridian Regional Airport) and Starkville (George M. Bryan Airport). The data was then subset to the following relevant hourly features: dewpoint temperature, dry bulb temperature, relative humidity, sea level pressure, visibility, wet bulb temperature, wind direction, and wind speed. Wind speed is our feature of interest, as wind speed is a critical factor when considering wind power. We further subset the data into three different months across 2022- January, July, and October. We chose these three months to represent diverse conditions as January usually implies winter weather, July implies summer weather, and October implies more mild weather. Table 1 shows a brief summary of the data.

For our predictive models, we chose four different RNNs: stacked stateless LSTM, stacked stateful LSTM, stacked stateless GRU, and stacked stateful GRU. Stacked architecture means that each neural network will have multiple layers, and each of our models had 10 layers. A stateful model uses the previous hidden state's output as initializers for the next hidden state, which is usually good for sequential data where one state is dependent on the next. For all of our models, we used the tanh activation function, the Adam optimizer, and 100 epochs. We used 70% of each month for training, and 30% for testing. Our models were evaluated using RMSE and the loss function.

## IV. RESULTS

Our results show that the stacked stateless LTSM model performs best with both the Meridian and Starkville datasets for every month measured. The full results measured by RMSE are shown in table 2. The stateful LSTM model had the worst overall performance in the Starkville dataset, and the stateful GRU model had the worst overall performance in the Meridian dataset. This is not what we expected because stateful models should perform better on sequential data where states are generally dependent on the previous state. This generalization may not be applicable to our data as wind speed is not necessarily dependent on the previous hour. For the months we used, July predictions were the most accurate. This may not mean that July is the best month for predictions, as both July datasets had the most data overall. Figure 1 shows a graphical summary of the best model's predictions (stacked stateless LSTM). Figure 2 shows a graphical representation of our worst performing model.

Table 1

| | Model | January 2022 | | July 2022 | | October 2022 | |
|---|---|---|---|---|---|---|---|
| | | Train RMSE | Test RMSE | Train RMSE | Test RMSE | Train RMSE | Test RMSE |
| **Starkville** | Stateless LSTM | **0.15** | **0.19** | **0.07** | **0.09** | **0.14** | **0.32** |
| | Stateful LSTM | 0.91 | 1.15 | 2.73 | 3.09 | 1.03 | 3.02 |
| | Stateless GRU | 0.16 | 0.20 | 0.29 | 0.32 | 1.41 | 2.79 |
| | Stateful GRU | 1.01 | 0.71 | 0.53 | 1.29 | 1.04 | 1.41 |
| **Meridian** | Stateless LSTM | 0.27 | 0.21 | 0.06 | 0.13 | 0.15 | 0.16 |
| | Stateful LSTM | 1.10 | 0.48 | 0.18 | 0.27 | 0.29 | 0.3 |
| | Stateless GRU | 0.50 | 0.59 | 0.09 | 0.16 | 0.58 | 0.61 |
| | Stateful GRU | 0.62 | 0.94 | 0.64 | 0.75 | 1.03 | 0.77 |

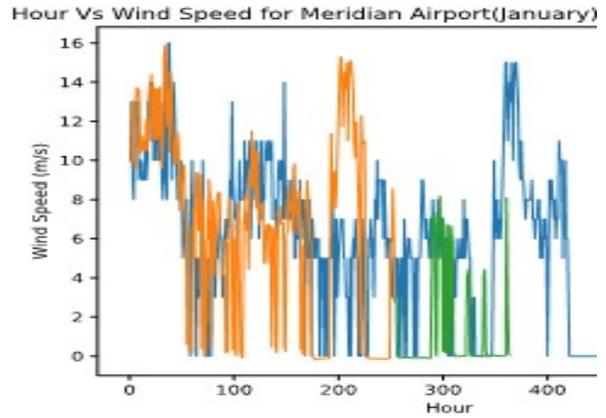
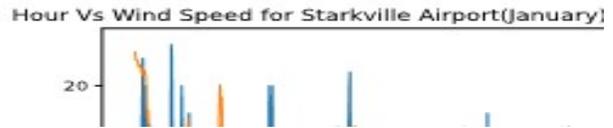

Fig 3 Starkville vs Meridian for Stacked Stateful LSTM

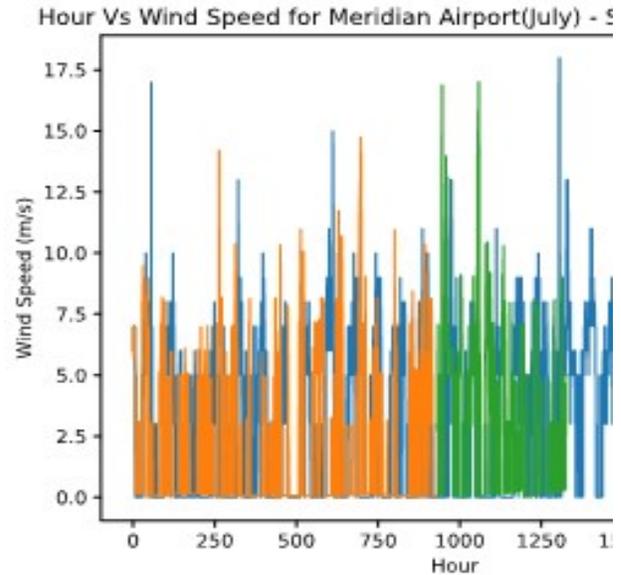
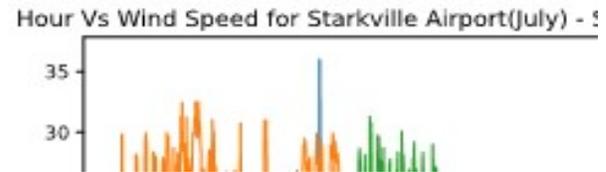

Fig 4 Meridian vs. Starkville for the month of July – Stacked Stateful LSTM

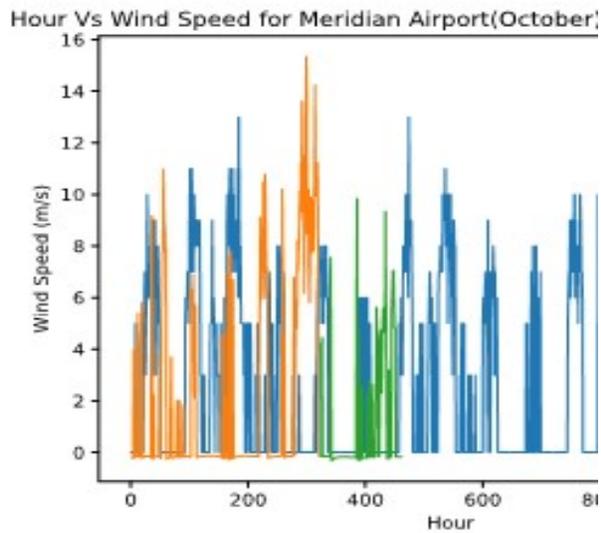
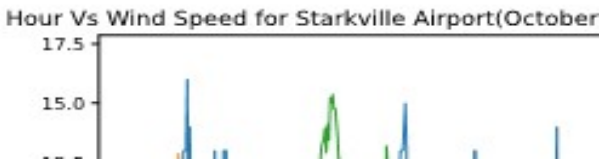

Fig 5 Meridian vs. Starkville for the month of October – Stacked Stateful LSTM

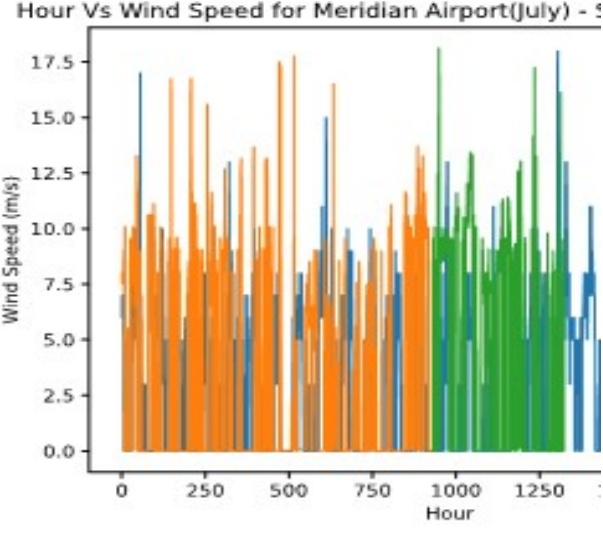
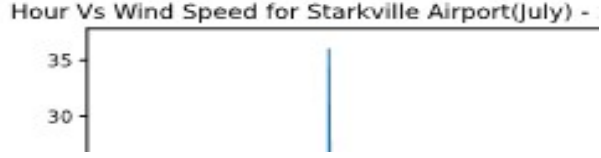

FIG 7 MERIDIAN VS. STARKVILLE FOR THE MONTH OF JULY – STACKED STATELESS LSTM

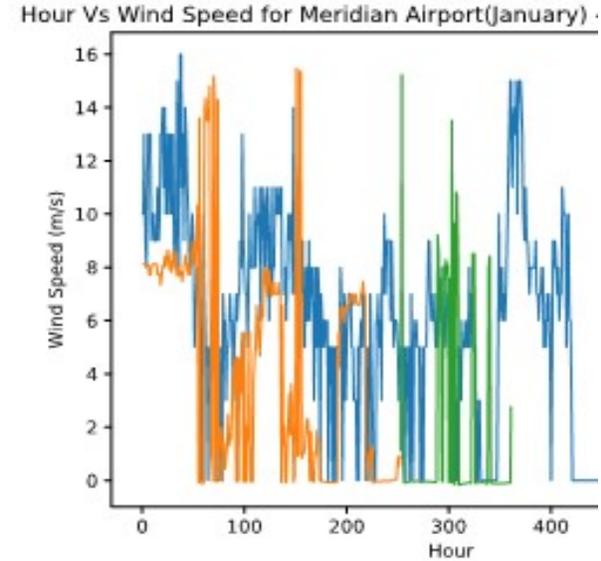
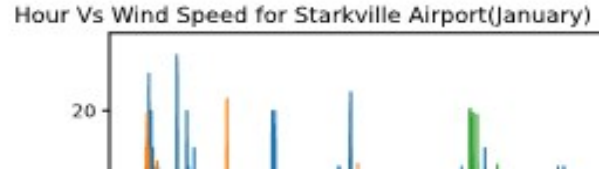

FIG 6 MERIDIAN VS. STARKVILLE FOR THE MONTH OF JANUARY – STACKED STATELESS LSTM

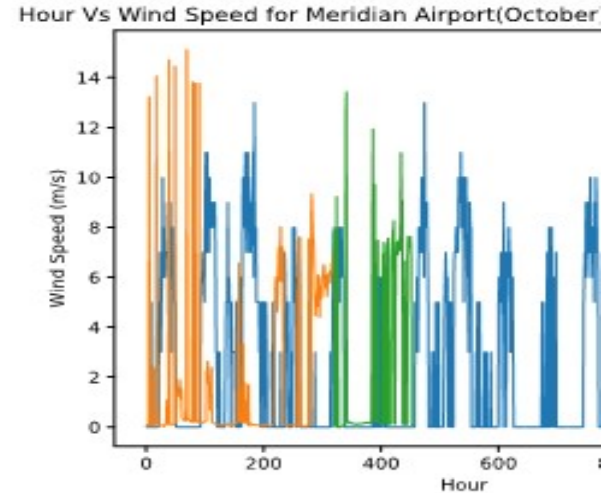
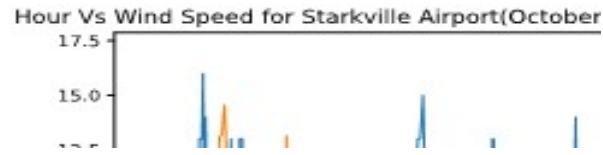

FIG 8 MERIDIAN VS. STARKVILLE FOR THE MONTH OF OCTOBER – STACKED STATELESS LSTM

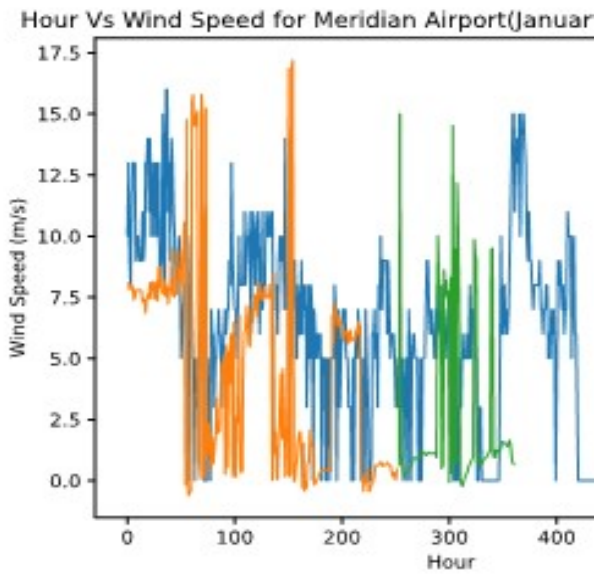
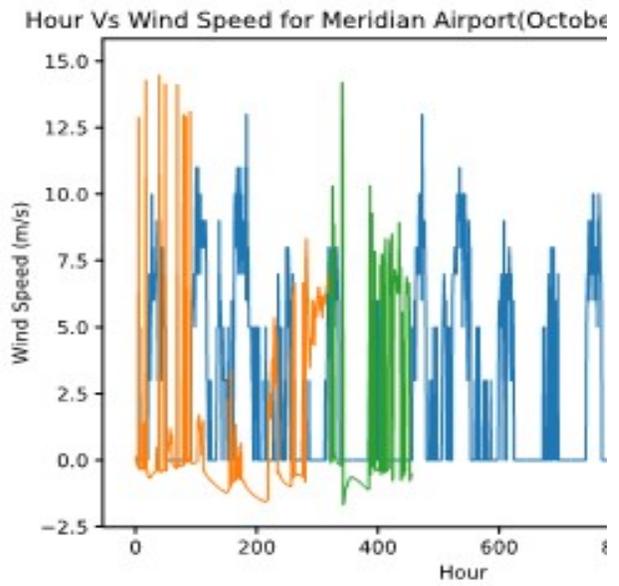
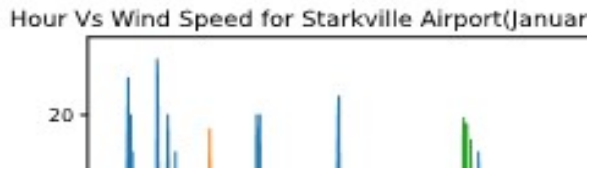
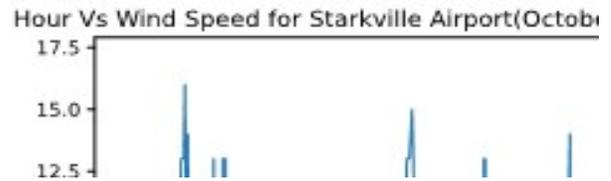

**FIG 9** MERIDIAN VS. STARKVILLE FOR THE MONTH OF JANUARY – STACKED STATEFUL GRU

**FIG 11** MERIDIAN VS. STARKVILLE FOR THE MONTH OF OCTOBER – STACKED STATEFUL GRU

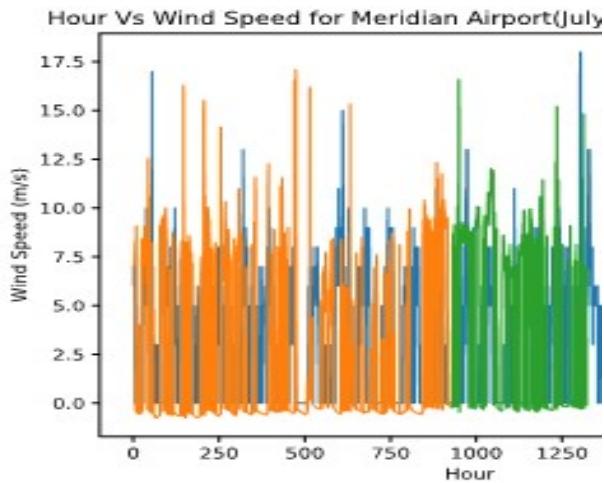
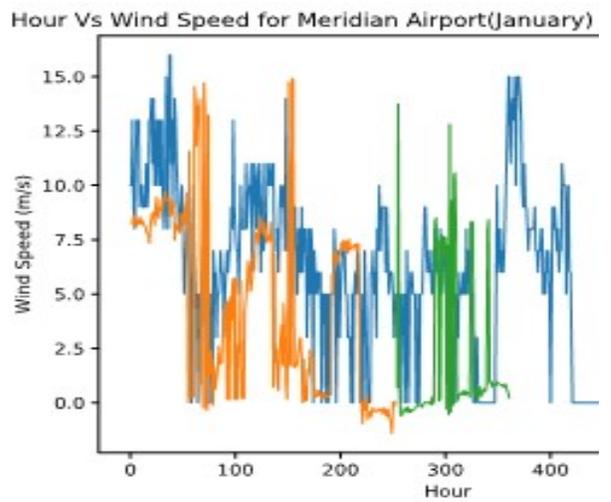
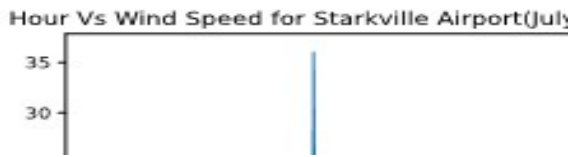
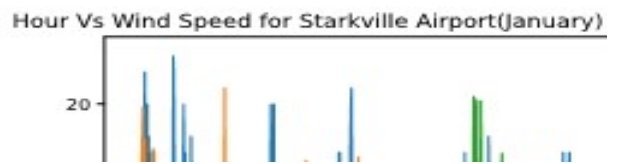

**FIG 10** MERIDIAN VS. STARKVILLE FOR THE MONTH OF JULY – STACKED STATEFUL GRU

**FIG 12** MERIDIAN VS. STARKVILLE FOR THE MONTH OF JANUARY – STACKED STATELESS GRU

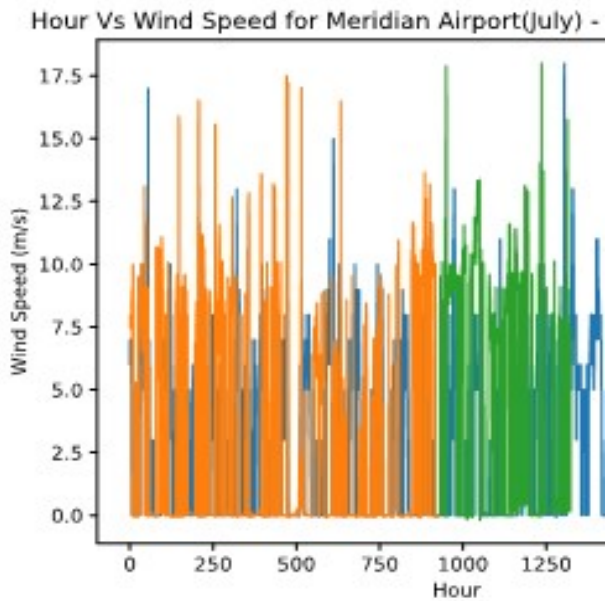

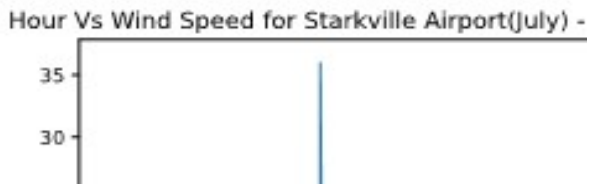

**FIG 13 MERIDIAN VS. STARKVILLE FOR THE MONTH OF JULY – STACKED STATELESS GRU**

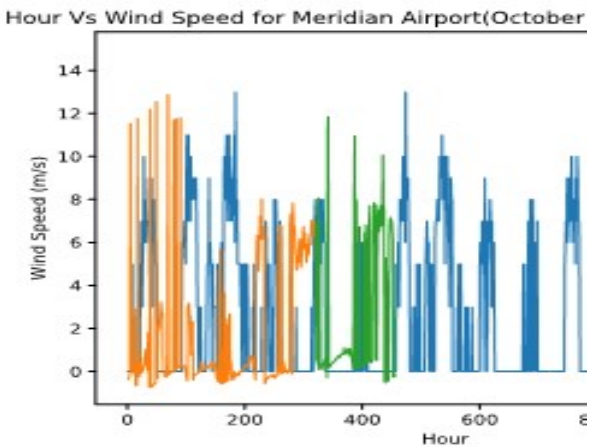

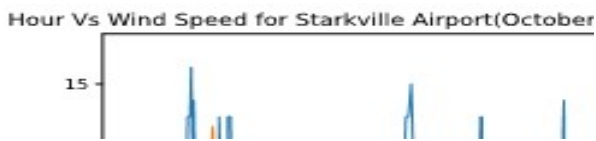

**FIG 14 MERIDIAN VS. STARKVILLE FOR THE MONTH OF OCTOBER – STACKED STATELESS GRU**

*Conclusion*

From the results it is clear that the stacked stateless LSTM had the best performance in both the Starkville and Meridian datasets, having the lowest RMSE in all tests. Typically, when dealing with time series data stateful models have higher predictive accuracy because each prediction is influenced by the one before. However, windspeeds are more affected by external factors than previous windspeeds. Out of the two locations chosen, Starkville had the higher average wind speed every month, however, in terms of practicality, neither location would be suitable for wind power as the wind speeds of both are simply too low. The aim of this paper was simply to test the methods for determining wind speeds rather than determining which location was better suited for wind turbines. However, the models used can be applied to other locations to accurately determine if they would be suitable. There are still several steps that can be taken to further the goals of this paper. Primarily, the models can be used over larger periods of time, and therefore, larger datasets. Hyperparameters could also be further tuned to give more accurate results. In addition, the models could theoretically be extended to make more long-term predictions of wind speed.

*Time Complexity*

The considered models use the following time complexities for its' execution:
Stacked Stateful LTSM (SSFLSTM) – $O(Tdh\log dh + Tdhdi)$
Stacked Stateless LTSM (SSLLSTM) - $O(Tdh\log dh + Tdhdi)$
Stacked Stateful GRU (SSFGRU) – $O(Td2h + Tdhdi)$
Stacked Stateless GRU (SSLGRU) - $O(Td2h + Tdhdi)$

*Future Directions*

We intend to compare results for at least four years to find which model performs best among the considered models. It would be good if we can implement the same considered models for long-term wind speed forecasting.

ACKNOWLEDGMENT

The research work would not have been possible if I did not get the immense guidance that I got from Dr. Pratik Chattopadhyay affiliated with Indian Institute of Technology Bhubaneshwar (IIT BHU) and Dr. Anindita Das Bhattacharjee affiliated with Maulana Abul Kalam Azad University of Technology, West Bengal, India.

CONTRIBUTION OF MEMBERS

Concept: Swayamjit Saha

Data Collection: Hunter Walt and Swayamjit Saha

Data Analysis: Hunter Walt

Paper writing: Ames Ugoyu, James Lucore, Swayamjit Saha and Hunter Walt